# Optimizing Odia Braille Literacy: The Influence of Speed on Error Reduction and Enhanced Comprehension

Monnie Parida
IIT Kharagpur
Kharagpur, West Bengal, India
mpiitkgp19@gmail.com

Manjira Sinha
TCS Research
Kolkata, West Bengal, India
manjiras@gmail.com

Anupam Basu
IIT Kharagpur
Kharagpur, West Bengal, India
anupambas@gmail.com

Pabitra Mitra
IIT Kharagpur
Kharagpur, West Bengal, India
pabitra@cse.iitkgp.ac.in

## ABSTRACT

This study aims to conduct an extensive detailed analysis of the Odia Braille reading comprehension among students with visual disability. Specifically, the study explores their reading speed and hand/finger movements. The study also aims to investigate any comprehension difficulties and reading errors they may encounter. Six students from the 9th and 10th grades, aged between 14 and 16, participated in the study. We observed participants' hand movements to understand how reading errors were connected to hand movements and identify the students' reading difficulties. We also evaluated the participants' Odia Braille reading skills, including their reading speed (in words per minute), errors, and comprehension. The average speed of Odia Braille reader is 17.64wpm. According to the study, there was a noticeable correlation between reading speed and reading errors. As reading speed decreased, the number of reading errors tended to increase. Moreover, the study established a link between reduced Braille reading errors and improved reading comprehension. In contrast, the study found that better comprehension was associated with increased reading speed. The researchers concluded with some interesting findings about preferred Braille reading patterns. These findings have important theoretical, developmental, and methodological implications for instruction.

## 1 INTRODUCTION

Individuals with visual disability, who rely on touch to access information, actively use their real-life experiences to read Braille. Tactile learning encompasses a range of strategies and materials, such as Braille, tactile graphics, and 3D models. Our objective is to provide a first hand experience of Braille reading, explicitly focusing on the Odia Braille script. Braille is a system of raised dots used for reading and writing by individuals with visual disability. A unique combination of up to six dots, arranged in two columns of three dots, represents each letter in Braille. In addition, there are symbols for punctuation, numbers, and frequently used words. Odia is a language spoken in the state of Odisha in eastern peninsular India. As a regional language, it is used widely across the state of Odisha. In Odisha, educators introduce students with visual disabilities to Braille using Odia Braille scripts. Unlike European Braille scripts like English, Odia Braille script has diacritics, consonant conjunct, and ligatures, essential to any Indian language. According to [5], Braille reading involves the movement of fingers in a scanning motion that focuses on crucial tactile information in a small, high-resolution area. This scanning motion is vital to acquiring knowledge through Braille reading.

This paper aims to examine the factors associated with reading Odia Braille. The subsequent sections discuss findings and insights drawn from the analysis of data collected from students with visual disability who participated in Odia Braille reading comprehension. The data comprises details on their finger scanning techniques, reading proficiency, and understanding obtained from a Braille comprehension test conducted during the research project's pilot phase.

## 2 RESEARCH QUESTION

- What is the speed in wpm for Odia Braille reader?
- What is the correlation between wpm and reading error for Odia Braille readers?
- Overall participants performance in Odia Braille reading comprehension task.
- What is the correlation between reading comprehension and reading error for Odia Braille readers?
- What is the correlation between reading comprehension and reading speed in wpm for Odia Braille readers?

## 3 RESEARCH METHODOLOGY

We carried out a pilot research project in Odisha's special schools to assess the reading comprehension of students with visual disability. We recorded the students' finger movements while they read the Braille script out loud, and we analyzed their hand scanning patterns.

### 3.1 Participants

The experiment involved 6 participants, three males and three females, who were 9th and 10th grade students at the Special School for Blind and Deaf Laxmiposi in Baripada, Odisha. All participants could read Odia Braille. We informed them about the experiment in detail, and special teachers were also consulted for approval and consent. All experiments were conducted in the presence of a special educator to ensure compliance with ethical protocols. The Institute Ethical Committee approved the research with an ethical clearance certificate.

### 3.2 Materials and Apparatus

We chose one prose and one passage from the 9th and 10th standard Odia high school syllabus as reading materials for our study. We manually converted the print versions of these materials and had them validated and proofread by a special educator. We made sure to select materials that were familiar enough so that unfamiliar vocabulary wouldn't hinder the students' understanding but not so familiar that the students were reciting from memory. We selected texts from textbooks meant for students to ensure all participants could understand the passages. We instructed the participants to read the entire passage aloud in Odia Braille format.

### 3.3 Procedure

The individual went through the reading material and responded to the questions. While recording the participants, we captured their hand movements on video and audio. The phone camera focused solely on the hands of the student and the full width of the reading material. It was done to capture the movement of the hands as they transitioned from line to line and page to page while intentionally excluding the student's face.

## 4 DATA ANALYSIS

We analyzed 11 video segments, each capturing a student reading aloud. We viewed each segment at least three times using the VLC media player plugin to measure response time in milliseconds after removing noise. We used this approach to improve accuracy. Each participant's answers were weighted on a scale of 5, as validated by a special educator. We recorded errors based on hand reading movement and audio. We identified reading errors based on video and audio recordings of finger movement.

In addition to collecting video and audio data during our experimentation, we made several observations and analysed various data points using SPSS software. We analysed the time taken by each participant to complete each task. We identified the number of errors made by each participant during the experiment to determine common mistakes and difficult words.

### 4.1 Average Speed of Odia Braille Reader

The average speed of an Odia Braille reader refers to the rate at which they read and process Odia Braille text. We measure it in words per minute (wpm), and it is calculated by dividing the total number of Odia words read by the time it takes to read them. The lack of precise testing methods restricts our understanding of the factors that limit Braille reading speed [3]. The limited number of participants significantly impacts the quality of our analysis. To address this issue and improve the accuracy of our findings, we have opted to increase our data set by incorporating two paragraphs with diverse content. Each paragraph has 113 and 91 Odia words, respectively. This will provide more significant variation in our data and enable us to draw more robust conclusions.

**Reading error** = (Total number of reading error / Total number of words)* 100

**Braille comprehension** = Sum(marks obtained by participant for each answer)/Total weightage for all the answers

**Table 1: Braille reading comprehension**

| Participant | Gender | WPM | Reading Error | Comprehension |
|---|---|---|---|---|
| P1 | Female | 5 wpm | 35.398 | 48 |
| P2 | Female | 19 wpm | 26.548 | 84 |
| P3 | Female | 12 wpm | 23.008 | 84 |
| P4 | Male | 9 wpm | 55.752 | 52 |
| P5 | Male | 31 wpm | 10.619 | 76 |
| P6 | Male | 9 wpm | 30.088 | 72 |
| P7 | Male | 41 wpm | 6.593 | 92 |
| P8 | Female | 17 wpm | 48.351 | 80 |
| P9 | Male | 13 wpm | 48.351 | 48 |
| P10 | Female | 29 wpm | 25.274 | 84 |
| P11 | Male | 9 wpm | 25.274 | 100 |

The average reading speed for Odia Braille reader is 17.64 wpm, which is relatively low compared to the average reading speed of (16.62±11.61) wpm in [1] for English Braille reader. The average of (116-149)wpm for English Braille in [4]. Again mean Braille reading in wpm by [2] is 100 wpm. For expert Braille reader about (64.94-185.29) wpm in [6]. The average reading speed can vary depending on several factors, such as practice, economic and social conditions, the prevalent use of audio versions of reading materials instead of Braille, and the level of difficulty of a language. The below graph shows the participant in the increasing order of their reading speed in wpm.

### 4.2 Correlation Between Reading Speed and Error for Odia Braille Reader

In Table 1 the reading error is calculated in percentage of error made by each participant. Although a participant may make multiple errors while reading the same word, we have assumed a count of 1 for each word, regardless of the different type of errors made. To calculate the correlation between reading speed in wpm and reading error for Odia Braille reader we have the below assumption

H1: Our null hypothesis assume we have no significant correlation between reading speed in wpm and reading error.

A strong negative (inverse) correlation was discovered between reading speed and error using SPSS, based on the Pearson product correlation and it was statistically significant(r=-0.667, n=11, p=0.025) with reading speed explaining 44.4%(coefficient of determination) variation in reading error. Hence H1 is discarded. This shows that an decrease in reading speed would result in higher reading error.

**Table 2: Correlation analysis between reading speed and reading error**

|       | WPM    | Error |
|-------|--------|-------|
| WPM   | 1      |       |
| Error | -0.667 | 1     |

*correlation is significant at 0.05 level(2tailed)

We can conclude that when reading speed decreases for Braille readers, it can be due to various finger movements and hand characteristics. Based on our observations, some Braille readers slow their reading speed to avoid errors, which can lead to pausing or hesitating. Additionally, Odia Braille readers tend to experience increasing reading errors and slower reading speed due to the use of regression during Odia Braille reading.

### 4.3 Odia Braille Reading Comprehension Performance

**Table 3: Odia Braille reading comprehension correctness percentage**

| Reading Compre1 |         | Reading Compre2 |         |
|-----------------|---------|-----------------|---------|
| Questions       | Correct | Questions       | Correct |
| 1               | 83.333% | 1               | 84%     |
| 2               | 46.666% | 2               | 96%     |
| 3               | 66.666% | 3               | 96%     |
| 4               | 50%     | 4               | 56%     |
| 5               | 60%     | 5               | 68%     |

To evaluate the participants' performance in the reading comprehension task, We assigned a total weightage of 5 to each correct answer with the help of a special educator. Based on this, we assigned marks to each participant's response. We calculated the percentage of correct answers using the below formula

**Correct percentage** = Sum(Participant marks for the individual Q&A) / Sum of total weightage for given Q&A.

We found that 96% of participants answered the second and third question correctly. This high percentage could be due to the questions being easy to understand and answer. However, we need to analyze more data and conduct statistical analysis to identify areas where improvements can be made. We aim to come up with concrete solutions based on our findings to improve the reading comprehension skills of Odia Braille readers.

### 4.4 Correlation between Reading Comprehension and Reading Speed for Odia Braille Readers

Braille reading involves two important aspects: reading comprehension and reading speed. Reading comprehension is the ability to comprehend and interpret the meaning of the Braille text, while reading speed refers to how quickly a person can read and process Braille dots. However, it's important to note that reading speed alone is not always indicative of reading comprehension. Even if someone reads quickly, they may struggle to understand or retain the information they read. Therefore, going further in this study, we will also consider reading errors in relation to reading comprehension. Here our goal is to determine the correlation between Braille reading speed in words per minute (wpm) and Braille reading comprehension for Odia Braille readers. Our hypothesis is as follows:

H1: Our null hypothesis assume we have no significant correlation between reading speed in wpm and reading comprehension.

Pearson product correlation of reading speed and reading comprehension was found to be moderately positive and statistically significant(r=0.467,n=11, p=0.148). Hence H1 is discarded. This shows that their is a significant correlation between reading speed in wpm and reading comprehension.

**Table 4: Correlation analysis for reading speed and reading comprehension**

|        | WPM   | Compre |
|--------|-------|--------|
| WPM    | 1     |        |
| Compre | 0.467 | 1      |

From above analysis we can conclude increasing reading speed leads to better comprehension, but it's important to consider that this correlation can be influenced by various factors. For instance, the participants' prior knowledge of the topic may play a role. In this case, as all participants belong to the same grade, they presumably have the same level of knowledge. Additionally, the difficulty level of the text and each participant's cognitive processing abilities are important factors to take into account which we have not considered.

### 4.5 Correlation Between Reading Comprehension and Reading Error for Odia Braille Readers

we already discussed above the need of correlation between Braille reading error and reading comprehension. So to determine the correlation between reading comprehension and reading error for Odia Braille reader our hypothesis is

H2: Our null hypothesis assume we have no significant correlation between reading error and reading comprehension.

Pearson product correlation of reading error and reading comprehension was found to be strong and negatively(inverse) correlated and statistically significant with(r=-0.657,n=11, p=0.028). Hence H2 is discarded. This shows that an decrease in reading error would result in much better reading comprehension.

**Table 5: Correlation analysis for reading error and reading comprehension**

|        | Error  | Compre |
|--------|--------|--------|
| Error  | 1      |        |
| Compre | -0.657 | 1      |

*correlation is significant at 0.05 level(2tailed)

From the above observation we can conclude if the participant does less Braille reading error the reading comprehension will be better.

## 5 DISCUSSION

The average reading speed of Odia Braille readers is 17.64 words per minute, which is lower compared to the reading speed of Braille readers in European countries. However, the level of comprehension among Odia readers may vary depending on the difficulty and clarity of the questions asked. To further analyze this, more statistical data would be needed, which is not within the scope of this paper. From the correlation observed, it can be concluded that better reading speed with fewer errors leads to better comprehension. On the other hand, slower reading speed with more errors leads to lower comprehension levels among readers.

The above finding are not only important from the point of view of Indian Braille scripts but will be applicable to any Braille script with similar characteristics of diacritics and consonant conjunct which we assume is responsible for the difficulty level in Braille readers.

## 6 LIMITATION

The study focused on the Odia Braille, a form of Braille used for the Odia language spoken in the eastern Indian state of Odisha. However, the experiment was limited by a small sample size as the participants required proficiency in reading and writing in Odia Braille. Moreover, there was no control group, which makes it challenging to eliminate the potential effects of factors like cognitive aspect and testing on the study outcomes. The limited sample size makes it tricky to employ machine learning algorithms and pattern recognition effectively.

## ACKNOWLEDGMENTS

We would like to express our sincere gratitude to "Special School for Blind and Deaf Laxmiposi, Baripada, Odisha" for their generous support and cooperation in allowing us to conduct our experiment. Their unwavering trust in our research have been invaluable. We are immensely grateful to the school administration, faculty, and students for their participation and contributions.